\documentclass{article}

\usepackage{arxiv}

\usepackage[utf8]{inputenc} 
\usepackage[T1]{fontenc}    
\usepackage{hyperref}       
\usepackage{url}            
\usepackage{booktabs}       
\usepackage{amsfonts}       
\usepackage{nicefrac}       
\usepackage{microtype}      
\usepackage{lipsum}

\usepackage{framed,multirow}
\usepackage{graphicx, amsmath, amsfonts}

\usepackage{amssymb}
\usepackage{latexsym}

\usepackage{url}
\usepackage{xcolor}
\definecolor{newcolor}{rgb}{.8,.349,.1}

\usepackage{fancyhdr}
\title{An Online Learning Approach for Dengue Fever Classification }

\author{
  Siddharth Srivastava, Sumit Soman, Astha Rai \\
  Centre for Development of Advanced Computing Noida, India\\
  \texttt{\{siddharthsrivastava, sumitsoman, asthar\}@cdac.in} \\
}

\begin{document}
\maketitle
\thispagestyle{fancy}
\begin{abstract}
This paper introduces a novel approach for dengue fever classification based on online learning paradigms. The proposed approach is suitable for practical implementation as it enables learning using only a few training samples. With time, the proposed approach is capable of learning incrementally from the data collected without need for retraining the model or redeployment of the prediction engine. Additionally, we also provide a comprehensive evaluation of machine learning methods for prediction of dengue fever. The input to the proposed pipeline comprises of recorded patient symptoms and diagnostic investigations. Offline classifier models have been employed to obtain baseline scores to establish that the feature set is optimal for classification of dengue. The primary benefit of the online detection model presented in the paper is that it has been established to effectively identify patients with high likelihood of dengue disease, and experiments on scalability in terms of number of training and test samples validate the use of the proposed model. 

\end{abstract}

\section{Introduction}

The use of learning algorithms or models for disease detection and prediction has recently evoked substantial interest in the health informatics community as it can efficiently aid physicians in diagnosing diseases using data-driven models. These approaches are also useful for obtaining long-term occurrence trends and causative symptoms for various diseases by efficiently mining data from healthcare information systems. Such tasks would traditionally be intractable to execute by human intervention, however, the use of machine learning methods would provide credible results by efficient processing of available, annotated data.

The primary challenge in developing disease detection models is the availability of labelled data that has been validated by clinicians. It is essential that learning models for disease detection should be trained on adequate training samples in order to be able to generalize better for test samples. However, since obtaining a large amount of labelled training data is not always practically feasible, as it would take a long time to curate such a dataset, one can alternatively employ a learning model that is initially trained using a few labelled samples, but incrementally learns and updates the model as and when new labelled data is available. In other words, online learning models present a viable use-case in such situations. Our objective in this paper is to develop a viable online learning model for the task of dengue disease detection, which is a mosquito-borne tropical disease which is difficult to diagnose, and therefore leads to fatalities. We now present a brief survey of recent literature in this domain to familiarize the reader with the efforts taken by researchers in this direction, and motivate the context and novelty of our work.

Potts \textit{et al.} \cite{potts2010prediction} have used laboratory findings collected 72 hours after fever onset (including White Blood Cell (WBC) count, percent monocytes, platelet count and hematocrit) to classify severity of dengue disease among patients visiting two hospitals (one rural and one urban) in Thailand. Althouse \textit{et al.} \cite{althouse2011prediction} used data retrieved from internet search for dengue occurrences in Singapore and Bangkok during 2004-2011 and evaluated multiple learning models to determine incidence of dengue. A similar effort using a larger dataset retrieved from internet search across the countries of Bolivia, Brazil, India, Indonesia and Singapore was presented in the work by Chan \textit{et al.} \cite{chan2011using}. Spatial and temporal models for dengue prediction have also been presented in the works by Dom \textit{et al.} for Malaysia \cite{dom2013generating}, Luz \textit{et al.} for Brazil \cite{luz2008time}, Phung \textit{et al.} for Vietnam \cite{phung2015identification}, Rotela \textit{et al.} for Argentina \cite{rotela2007space} among others. 

On a global scale, Bhatt \textit{et al.} \cite{bhatt2013global} have provided analysis on a larger dataset using reported occurrences to develop a modeling framework for global dengue risk determination. Hales \textit{et al.} \cite{hales2002potential} determine the effect of global climate change on vector borne diseases, specifically based on vapor pressure levels. In fact, climate based dengue disease prediction has appeared in other works as well such as that of Descloux \textit{et al.} \cite{descloux2012climate}, Hii \textit{et al.} \cite{hii2012forecast}, Pinto \textit{et al.} \cite{pinto2011influence}, Earnest \textit{et al.} \cite{earnest2012meteorological}.

Multiple learning approaches have also been employed for dengue detection, including Decision-Tree algorithms \cite{tanner2008decision}, neural networks \cite{aburas2010dengue}, ensemble based methods \cite{loshini2015predicting}, Adaptive Neuro-Fuzzy Inference System \cite{faisal2012adaptive}, among others. A significant number of approaches in literature have employed Support Vector Machines and their variants for dengue disease prediction. These include works by Wu \textit{et al.} \cite{wu2008detect}, Yusof \textit{et al.} \cite{yusof2011dengue}, Khan \textit{et al.} \cite{khan2016analysis}, Gomes \textit{et al.} \cite{gomes2010classification}, Radzol \textit{et al.} \cite{radzol2014classification} among others.

As can be seen, most efforts in literature have performed offline analysis on dengue datasets which have been reported, and it is evident that the detection of this disease from clinical datasets is of use to the clinicians as well as healthcare policy makers. In order to practically realize the benefits of a dengue detection system, it is important to have the model deployed in a healthcare information system where it can analyze data samples in real-time and flag samples with high likelihood of the disease to the users of the system. These can then aid the clinician to arrive at a diagnosis for specific patients. Moreover, it can be useful to data scientists and clinical researchers in identifying and evaluating the factors that contribute to efficient detection of dengue. In this paper, we obtain a novel dataset derived from a healthcare information system deployed across multiple healthcare facilities. The dataset is pre-processed to obtain a feature space which has a combination of patient symptoms and clinical investigations that are conventionally used to diagnose dengue. We first establish the adequacy of this feature space for dengue detection by evaluating a few classifier models and analyzing their scalability by varying the number of training and test samples. Subsequently, we proceed to develop an online dengue classification model which is trained in an online manner (the classifier models is updated as and when new samples with dengue as the diagnoses are validated by the physician). To this end, we evaluate multiple online classifier models and provide a comprehensive report of their performance. Our results demonstrate that the obtained feature representation and online classifier trained can be practically used in a healthcare information system for dengue detection.

The rest of the paper is organized as follows. Section \ref{sec:system} discusses the design and architecture of an online model for dengue classification that can be used in a healthcare information system. A discussion of the machine learning methods that have been analyzed on the dataset in this paper is presented in Section \ref{sec:methods}. Section \ref{sec:experiments} comprises of a description of the features used in our dataset along with results using the classifier models. Finally, conclusions and future work are presented in Section \ref{sec:conclusions}.

\section{Methodology and System Design \label{sec:system}}

\begin{figure}[hbtp]
    \centering
    \includegraphics[scale=0.60]{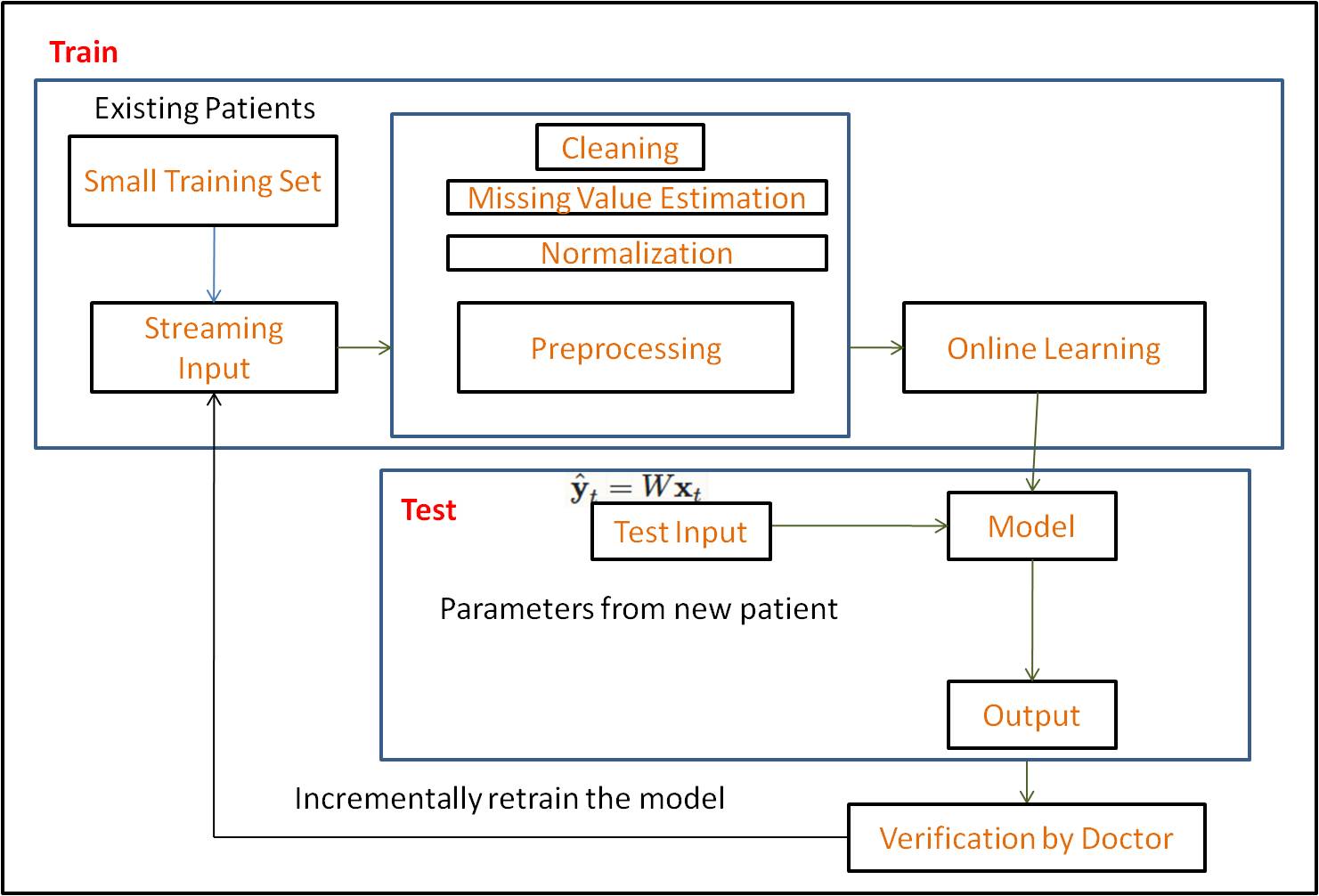}
    \caption{Dengue Detection System Architecture}
    \label{fig:flow}
\end{figure}

We begin with a description of the design and architecture of the dengue classification system, shown in Fig. \ref{fig:flow}, that can be integrated into a healthcare information system for online predictions on streaming data samples. The pre-requisite for using such a model is that the system should record clinical data, including symptoms and results of laboratory investigations, for patients along with a suitable mechanism for recording diagnoses for a patient, which is mapped to patients via suitable identifiers \cite{soman2015unique, srivastava2016high}. Once such a system is in place, a dataset generation and model training component can be incorporated into the healthcare information system. However, for this model to learn incrementally from new data samples, a diagnoses validation user interface also needs to be provided, which can validate the clinician's diagnoses with that arrived at by the learning model, in order to incrementally improve the learning model to deliver robust predictions.

A two-phase approach has been adopted in this work to develop such an online dengue prediction model. Initially, we extract the features from a small training set which comprises of annotated samples of patients diagnosed with dengue fever. The dataset is derived after multiple pre-processing steps which involve extracting the data samples from the relational database used by the healthcare information system, cleaning the dataset to extract the features, which may also include steps to estimate missing values as well as normalize the dataset. Using these small number of samples represented in an appropriate feature space, we train a classifier model to predict whether a patient is likely to be diagnosed with dengue, based on the symptoms presented that have been used to train the model. 

It may be noted here that choosing the feature space in which the data samples are to be represented is also a critical task as the generalization ability of the model depends on the diversity of the chosen features. It is important to choose as many features as possible that could be significant in detecting dengue, while also keeping in mind that the features should be such that there should be high availability in the healthcare information system (which means that it should be pertinently recorded across samples), while also being accurate in terms of precision and representation formats. In an ideal scenario, symptoms, investigations and diagnoses should be recorded using healthcare standards such as SNOMED \cite{donnelly2006snomed}, LOINC \cite{mcdonald2003loinc}, among others \cite{srivastava2014electronic}. However, the integration of such standards into legacy hospital information systems is still in its infancy \cite{srivastava2017continuity}. Therefore, in our work, we choose a set of features that are primarily categorical in nature and encompass the broad set of symptoms and laboratory investigations mandated for diagnosing dengue. This enables easier extraction and generation of data samples for classifier training, while also reducing the complexity of the dengue detection model. Moreover, the nature of features chosen makes the model ubiquitous for implementation across multiple healthcare information systems as well.

In the next phase of the dengue classification system, the trained classifier is used to obtain predictions on unseen samples (new patients whose symptoms and investigation results have been recorded and likelihood for dengue as a diagnosis needs to be evaluated). The predictions arrived at by the online learning model are then validated by clinicians. These labeled data samples are then used to re-train the classifier, in order to improve its generalization. Over a period of time, as more and more training data becomes available, the classifier training will become more robust and high accuracy predictions would be delivered by the dengue classification system.

In the following section, we discuss the learning models on which our dataset has been evaluated, which includes algorithms for offline and online classifier training.

\section{Learning Algorithms Evaluated\label{sec:methods}}
In this section, we briefly describe the learning algorithms used to evaluate the generalization on the dataset. The idea is to be able to comprehensively evaluate how multiple learning online algorithms perform so that an optimal learning model could be used for online evaluation in applied systems. 

A brief of the notation used in the following sub-sections is discussed for the convenience of the reader. We shall denote a data sample in $n$ dimensions by $x \in \mathbb{R}^n$. The data matrix constitutes of $M$ such samples and is denoted by $X \in \mathbb{R}^{M \times n}$. The labels for the samples are indicative of the class to which the sample belongs to, and for the case of binary classification (with two classes), is denoted as $Y=\{y^i|y^i \in \{+1,-1\}, \forall i=1,2,...,M\}$. The weight vector, which represents the decision boundary for identifying the class of a sample is denoted by $w \in \mathbb{R}^n$, and may also constitute of a bias term denoted by $b \in \mathbb{R}$. The weight vector is conventionally distributed normally as $\mathbb{N}(\mu,\Sigma)$, where $\mu$ and $\Sigma$ denote the mean and co-variance of the distribution respectively.

\subsection{Offline Learning Algorithms}

These algorithms have primarily been used to establish the adequacy of the feature set for the task of dengue disease classification. Baseline results have been obtained using the Support Vector Machine (SVM) and Random Forests, which have been discussed below.

\subsubsection{Support Vector Machine (SVM)}

The Support Vector Machine (SVM) \cite{cortes1995support} for binary classification solves the following Quadratic Programming Problem (QPP).
\begin{gather}
\min_{w,b,\xi} \frac{1}{2} \|w\|_2^2 + C \sum_{i=1}^M \xi_i \label{svm_primal_obj}\\
\text{subject to,} \nonumber \\
y^i(w^T x^i + b) + \xi_i \geq 1,  \forall i = 1,2,...,M \label{svm_primal_const1}\\
\xi_i \geq 0, \forall i = 1,2,...,M. \label{svm_primal_const2}
\end{gather}

Here, the separating hyperplane is $w^T x  + b = 0$ and $\xi_i$ are the positively constrained slack variables. $C$ is a hyperparameter that controls the tradeoff between minimizing the $L_2$ norm of the weight vector and the slack variables. The class to which a test point $x \in \mathbb{R}^n$ belongs is found by evaluating $sgn(w^T x + b)$. We mention here in passing that the SVM is largely popular owing to the kernel trick which allows mapping the input features to a high-dimensional space where a linear classifier can be found, and that the dual formulation of (\ref{svm_primal_obj})-(\ref{svm_primal_const2}) is practically used owing to its tractability. 

\subsubsection{Random Forests}

Random forests, or random decision trees \cite{ho1995random} consist of generating multiple decision trees and then using their combination to arrive at a prediction on a test sample. Each decision tree is trained on a subset of the training samples drawn with replacement from the training dataset. Random forests are also trained using a subset of features of the dataset instead of the entire dataset. Multiple variants of this method are also available in literature. 

\subsection{Online Learning Algorithms}

We evaluate multiple online learning algorithms, which are briefly summarized in the following subsections. The general working mechanism of the online learning algorithms is as follows. Each algorithm suitably initializes the model parameters $w$ and $b$ and stochastically updates these parameters based on different update rules. These update rules are often dependent on evaluation of a suitable loss function, which is computed by the predicted label and the true label of the incoming sample. It is expected that the training will converge to the optimal classifier model parameters as it is trained stochastically.

\subsubsection{Adaptive Regularization Of Weights (AROW) and its variants}

The Adaptive Regularization Of Weights (AROW) \cite{crammer2009adaptive} uses adaptive regularization on the weight vector with each training sample in order to make the learning model robust to make it robust to label noise. The algorithm begins by initializing $w=0$ and $\Sigma=I$, where $I$ denotes the identity matrix. For each sample, the weight vector is initially updated by the mean $\mu^i$ and the prediction on the sample is computed as $sgn(w^T x^i)$. Based on the true label $y^i$, the loss is computed using (\ref{eqn:arow1}), and the model parameters are updated as given by (\ref{eqn:arow2})-(\ref{eqn:arow3}) when the computed loss term is positive.

\begin{gather}
l_i=\max\{0,(1-y^i w^T x^i)\} \label{eqn:arow1} \\
\mu_{i+1}=\mu_i + \alpha_i \sum_i y^i x^i \label{eqn:arow2} \\
\Sigma_{i+1}=\Sigma_i + \beta_i \sum_i x^i {x^i}^T \Sigma_i \label{eqn:arow3} 
\end{gather}
\noindent where $\alpha_i = l_i(\mu_i,(x^i,y^i))\beta_i$ and $\beta_i=\frac{1}{{x^i}^T \Sigma_i x^i + \gamma}$ for a suitable hyper-parameter $\gamma$.

A variant of the AROW, the New Adaptive Regularization of Weights (NAROW) uses another additional second order term which restricts the eigenvalues of $\Sigma$ within specific bounds. The update rules for NAROW are given by (\ref{eqn:narow1})-(\ref{eqn:narow4}), where $b > 0$ is a suitably chosen bound.

\begin{gather}
    \mu_{i+1}=\mu_i + y^i x^i \label{eqn:narow1} \\
    \Sigma_{i+1}=\Sigma_i + \beta_i \sum_i x^i {x^i}^T \Sigma_i \label{eqn:narow2} \\
    \text{where, }\beta_i=\frac{1}{{x^i}^T \Sigma_i x^i + \gamma_i} \label{eqn:narow3} \\
    \gamma_t=\begin{cases}
    \frac{{x^i}^T \Sigma_i x^i}{b {x^i}^T \Sigma_i x^i - 1}, & {x^i}^T \Sigma_i x^i \geq \frac{1}{b} \\
    +\infty, & \text{otherwise} 
    \end{cases} \label{eqn:narow4}
\end{gather}

\subsubsection{Online Gradient Descent}

The Online Gradient Descent (OGD) \cite{zinkevich2003online} is a stochastic or batch-wise update version of the classical gradient descent algorithm for optimization. The approach is similar to the AROW except for the weight update rule, which is given by (\ref{eqn:ogd_update}) for OGD.

\begin{gather}
w_{i+1}=w_i + l_i \sqrt{\frac{1}{i y^i x^i}} \label{eqn:ogd_update}
\end{gather}

In principle, the OGD combines conventional gradient descent with the hinge loss function. The stochastic update rules makes it amenable for large datasets.

\subsubsection{Confidence-weighted learning (CW) and Soft variants (SCW 1 \& SCW 2)}

Confidence-weighted learning aims at updating the weight distribution using the Kullback-Leibler divergence ($D_{KL}$) with the constraint that the probability of correct classification is constrained by a determined threshold $\eta$. This involves the update rule given by the optimization problem in (\ref{eqn:cw1})-(\ref{eqn:cw2}).

\begin{gather}
(\mu_{i+1},\Sigma_{i+1}) = \arg \min_{\mu,\Sigma} D_{KL}(\mathbb{N}({\mu,\Sigma})\| \mathbb{N}({\mu_i,\Sigma_i})) \label{eqn:cw1} \\
\text{subject to} \nonumber \\
P_{w \in \mathbb{N}({\mu,\Sigma}) }(y^i [w^T x^i] \geq 0) \geq \eta \label{eqn:cw2}
\end{gather}

The SCW improves upon the CW and AROW by the use of a modified loss function which is given by (\ref{eqn:scw1}) for SCW 1 and (\ref{eqn:scw2}) for SCW 2.

\begin{gather}
(\mu_{i+1},\Sigma_{i+1}) = \arg \min_{\mu,\Sigma} D_{KL}(\mathbb{N}({\mu,\Sigma})\| \mathbb{N}({\mu_i,\Sigma_i})) + \nonumber \\
C l^\phi (\mathbb{N}(({\mu,\Sigma});(x^i,y^i))) \label{eqn:scw1} \\
(\mu_{i+1},\Sigma_{i+1}) = \arg \min_{\mu,\Sigma} D_{KL}(\mathbb{N}({\mu,\Sigma})\| \mathbb{N}({\mu_i,\Sigma_i})) + \nonumber \\ C l^\phi (\mathbb{N}(({\mu,\Sigma});(x^i,y^i)))^2 \label{eqn:scw2} 
\end{gather}

\noindent where $C$ is a hyper-parameter and $l^\phi (\mathbb{N}(({\mu,\Sigma});(x^i,y^i))=\max(0,\phi \sqrt{{x^i}^T \Sigma x^i} - y^i \mu x^i)$. Here, $C$ is a hyperparameter that determines the relative weight to be assigned to the terms in the update rule.

\subsubsection{Normalized HERD (NHERD)}

The Normalized HERD (NHERD) \cite{crammer2010learning} uses the loss function defined by (\ref{eqn:nherd1})-(\ref{eqn:nherd3}), in order to herd a normalized weight distribution by constraining the velocity flow. It may be noted here that $\mathbb{E}$ denotes the expectation operator. 

\begin{gather}
\mu_{i+1}=A_i \mu_i + b_i \label{eqn:nherd1} \\
\Sigma_{i+1}=A_i \Sigma_i A_i^T \label{eqn:nherd2} \\
\text{where, }(A_i,b_i)=\arg \min_{A,b} \mathbb{E}_{w \in \mathbb{N}(\mu_i,\Sigma_i)} C_i(A_i w^i + b_i) \label{eqn:nherd3} \\
C_i (w)=\frac{1}{2}[{(w-w^i)}^T \Sigma_i^{-1} (w-w^i) + C \max {(0,1-y^i w^T x^i)}^2
\end{gather}

\subsubsection{PA and its variants PA1 \& PA2}

Passive Aggressive (PA) \cite{crammer2006online} learning methods update the weight of the classifier model in the current iteration by minimizing the loss suffered by the classifier on the current sample. This effectively minimizes the error obtained by the classifier model trained in the previous iteration. The update rules are given by (\ref{eqn:pa})-(\ref{eqn:pa1a}).

\begin{gather}
w_{i+1}=\arg \min_w \frac{1}{2} {\|w-w^i\|}^2 \label{eqn:pa} \\
\text{subject to, } \max(0,1-y^i w^T x^i)=0 \label{eqn:pa1a}
\end{gather}

Extensions to PA include PA1 and PA2, whose objective functions are given by (\ref{eqn:pa1}) and (\ref{eqn:pa2}) respectively. 

\begin{gather}
w_{i+1}=\arg \min_w \frac{1}{2} {\|w-w^i\|}^2 + C \max(0,1-y^i w^T x^i) \label{eqn:pa1} \\
w_{i+1}=\arg \min_w \frac{1}{2} {\|w-w^i\|}^2 + C {\max(0,1-y^i w^T x^i)}^2 \label{eqn:pa2}
\end{gather}

Here, $C > 0$ is a suitably chosen hyperparameter to control the weight assigned to the individual terms in the update rule. It can be seen that PA1 is an unconstrained version of PA which can directly be solved by gradient descent based methods, and hence is suitable for larger datasets. PA2 is similar to PA1, except that it uses the squared loss function.

\subsubsection{Improved Ellipsoid Method (IELLIP)}

The Improved Ellipsoid Method (IELLIP) \cite{yang2009online} uses the update rules as given by (\ref{eqn:iellip1})-(\ref{eqn:iellip4}).

\begin{gather}
\mu_{i+1}=\mu_i + \alpha_i \sum_i g_i \label{eqn:iellip1}\\
\Sigma_{i+1}=\frac{1}{1-c_i}(\Sigma_{i} c_i \sum_i g_i g_i^T \Sigma_i \label{eqn:iellip2} \\
\text{where, } \alpha_i =\frac{\alpha \gamma - y^i \mu_i^T x^i}{\sqrt{{x^i}^T\Sigma_i x^i}} \label{eqn:iellip3} \\
g_i=\frac{y^i x^i}{\sqrt{{x^i}^T\Sigma_i x^i}}, c_i=c b^T, 0 \leq c,b \leq 1 \label{eqn:iellip4}
\end{gather}

\subsubsection{Approximate Large Margin Algorithm (ALMA)}

The Approximate Large Margin Algorithm (ALMA) \cite{gentile2001new} implements an alternative loss function characterized by parameter $\alpha$ in order to obtain a classifier model with margin dependent on the parameter. The update rule is given by (\ref{eqn:alma1})-(\ref{eqn:alma2}).

\begin{gather}
w=\frac{w+l_i \sqrt{\frac{2}{k}} y^i \frac{x^i}{\|x^i\|}}{\max(1,\|w+l_i \sqrt{\frac{2}{k}} y^i \frac{x^i}{\|x^i\|}\|)} \label{eqn:alma1}\\
\text{where, }l_i=\max(0,\frac{1-\alpha}{\alpha\sqrt{k}}-\frac{y^i w^T x^i}{\|x^i\|}), k=k+l_i \label{eqn:alma2}
\end{gather}

\subsubsection{Perceptron and Second Order Perceptron (SOP)}
The classical perceptron \cite{rosenblatt1958perceptron} learning model is given by (\ref{eqn:perceptron1})-(\ref{eqn:perceptron2}). It simply updates the classifier weights by a loss function dependent upon the number of misclassified samples.

\begin{gather}
w^{i+1}=w^i+l_i y^i x^i \label{eqn:perceptron1} \\
\text{where, } l_i=\sum_i(y^i_{pred}\neq y^i) \label{eqn:perceptron2}
\end{gather}

\noindent Here, $y^i_{pred}$ denotes the predicted label on the sample $x^i$ and $y^i$ is the true label of the sample. A variant of the perceptron, the Second Order Perceptron (SOP) \cite{cesa2005second}, works in a similar manner while also additionally updating the covariance matrix $\Sigma$ by a factor of $\alpha I$, where $\alpha>0$ is a suitably chosen hyperparameter and $I$ is the identity matrix.

\subsection{Relaxed Online Maximum Margin Algorithm (ROMMA) \& aggressive ROMMA (aROMMA)}

The Relaxed Online Maximum Margin Algorithm (ROMMA) \cite{li2000relaxed} uses an alternate weight update rule as given by (\ref{eqn:romma1}).

\begin{gather}
w_{i+1}=\frac{{\|x^i\|}^2 {\|w^i\|}^2 - y^i w^T x^i}{{\|x^i\|}^2 {\|w^i\|}^2 - {(w^T x^i)}^2}w_i + \frac{{\|w^i\|}^2(y^i - w^T x^i)}{{\|x^i\|}^2 {\|w^i\|}^2 - {(w^T x^i)}^2}x^i \label{eqn:romma1}
\end{gather}

This weight update is conditional to the evaluation of a loss function $l_i$ (defined by (\ref{eqn:romma2})) returning a positive value.

\begin{gather}
l_i=\begin{cases}
\sum_i y^i_{pred}\neq y^i & \text{for ROMMA}, \\
\sum_i (\max(0,1-y^i w^t x^i)>0) & \text{for aROMMA}
\end{cases} \label{eqn:romma2}
\end{gather}

In the following section, we extensively evaluate the offline and online learning algorithms for the task of dengue disease detection. As mentioned previously, the baseline results on the dataset is obtained using the offline classifiers. Since the actual system is would need to work on streaming data, the online classifier models are vital in building a robust dengue classification system.

\section{Experiments and Results\label{sec:experiments}}

This section describes our experimental setup and results obtained. We first describe the features available in our dataset. Our initial analysis is presented in an offline setting, where we initially use Support Vector Machines (SVMs), Random Forests, as well as multiple classifiers from the Weka Machine Learning library. Subsequently, we evaluate classifiers in an online setting, which establishes the use of our proposed pipeline for detection from streaming data.

\subsection{Dataset Description}

We extracted the following features, which comprise of symptoms and results of pathological investigations, from the dataset of patients in the healthcare information system:
\begin{enumerate}
    \item No. of days for which symptoms appeared (numeric)
    \item Vomiting/Nausea as a symptom (1 if yes, 0 if no)
    \item Severe frontal headache (1 if yes, 0 if no)
    \item Retro Orbital Pain  (1 if yes, 0 if no)
    \item Rash  (1 if yes, 0 if no)
    \item Abdominal Pain  (1 if yes, 0 if no)
    \item Muscle/Bone Pain  (1 if yes, 0 if no)
    \item High-grade fever or pyrexia  (1 if yes, 0 if no)
    \item Hemorrhagic Manifestation  (1 if yes, 0 if no)
    \item Non-structural glycoprotein-1 (NS1) Antigen
    \item Dengue Virus IgM Enzyme-Linked ImmunoSorbent Assay (ELISA)
    \item Serum Immunoglobulin IgG ELISA
    \item Dengue IgM ELISA 1
\end{enumerate}

The label to be predicted (diagnosis if dengue is positive or not) was represented as the final categorical variable in the dataset. Of the total 182 samples present in the processed dataset, 148 had positive dengue diagnosis and the remaining 34 were diagnosed negative.

\subsection{Baseline Results using Offline Classifiers}

The first step in developing the model was to ascertain that the feature representation for the data samples was adequate for dengue detection. For this task, we evaluated the classification accuracy on the dataset using SVM classifier for multiple train-test split ratios, as shown in Table \ref{res:svm}. It can be seen that though the initial classification accuracy is low, it improves as the classifier model is trained with a larger number of training samples. It is to be noted that the training of the SVM model in this case is not stochastic (or online), rather the entire training dataset is used to train the classifier in one go. One can observe that the highest classification accuracy obtained is \textbf{85.29\%}, which is obtained when using 80\% of the data samples for training and remaining 20\% for testing.

\begin{table}[!t]
\centering
\caption{Results using SVM with varying train-test split}
\label{res:svm}
\scalebox{1.0}{
\begin{tabular}{|c|c|}
\hline
\multicolumn{1}{|l|}{\textbf{Train:Test}} & \multicolumn{1}{l|}{\textbf{Accuracy(\%)}} \\ \hline
10:90                                     & 61.34                                  \\ \hline
20:80                                     & 65.27                                  \\ \hline
30:70                                     & 77.95                                  \\ \hline
40:60                                     & 82.24                                 \\ \hline
50:50                                     & 78.89                                  \\ \hline
60:40                                     & 72.22                                  \\ \hline
70:30                                     & 84.90                                  \\ \hline
80:20                                     & \textbf{85.29}                                  \\ \hline
90:10                                     & 82.35                                  \\ \hline
\end{tabular}%
}
\end{table}

The results using random forest classifier with varying the number of decision trees is shown in Table \ref{res:rf}. It may be noted that the train-test ratio for these results is kept at 80-20. The highest classification accuracy obtained using random forests for our dataset is \textbf{98.13\%}.

\begin{table*}[]
\centering
\caption{Results on Random Forest with varying decision trees}
\label{res:rf}
\scalebox{1.0}{
\begin{tabular}{|l|l|l|l|l|l|l|l|l|l|}
\hline
\#Decision trees  & 1              & 2     & 3     & 4      & 5      & 6      & 7      & 8      & 9      \\ \hline
Accuracy          & 97.92          & 97.86 & 97.97 & 97.64  & 97.97  & 98.03  & 97.97  & 98.3   & 97.53  \\ \hline
\# Decision trees & 10             & 20    & 30    & 40     & 50     & 60     & 70     & 80     & 90     \\ \hline
Accuracy          & \textbf{98.13} & 87.8  & 88.08 & 87.852 & 87.472 & 87.638 & 87.637 & 87.637 & 87.967 \\ \hline
\end{tabular}}
\end{table*}

These results establish that the chosen feature set is suitable for the task of dengue disease classification. The improvement in accuracy when using larger amounts of training data leads us to develop an online learning model that can adaptively improve as more and more labeled data for newer patients becomes available to the dengue detection system. In the present context, we use our dataset with multiple online classifiers to present the viability of the proposed dataset and learning model.

\subsection{Results using multiple classifiers}

Table \ref{tab:weka} shows results on the dataset using various classification algorithms from the Weka library, on the performance metrics of accuracy, true positive and false positive rate, precision, recall, f-measure and Mean Classification Accuracy (MCC). The metrics other than accuracy are often employed in evaluating classifiers on imbalanced data, where the distribution of samples between the classes is skewed. We find that most algorithms perform similarly in terms of accuracy, while random forests outperform other classifiers with an accuracy of 82.41\%. Interestingly, the false positive rate in case of random forests is nearly half than that of the other classifiers. This observation is significant in applications such as disease detection where avoiding incorrect diagnosis is of significant.

\begin{table*}[hbtp]
\centering
\caption{Results using classification algorithms}
\label{tab:weka}
\begin{tabular}{|l|l|l|l|l|l|l|l|l|}
\hline
\multicolumn{1}{|c|}{\textbf{S.No.}} & \multicolumn{1}{c|}{\textbf{Algorithm}} & \multicolumn{1}{c|}{\textbf{Accuracy}} & \multicolumn{1}{c|}{\textbf{TP Rate}} & \multicolumn{1}{c|}{\textbf{FP Rate}} & \multicolumn{1}{c|}{\textbf{Precision}} & \multicolumn{1}{c|}{\textbf{Recall}} & \multicolumn{1}{c|}{\textbf{F-Measure}} & \multicolumn{1}{c|}{\textbf{MCC}} \\ \hline
1                                    & Logistic                                & 78.02                                 & 0.780                                 & 0.753                                 & 0.712                                   & 0.780                                & 0.735                                   & 0.043                             \\ \hline
2                                    & SGD                                     & 81.31                                  & 0.813                                 & 0.813                                 &                                         & 0.813                                &                                         &                                   \\ \hline
3                                    & SGD Text                                & 81.31                                  & 0.813                                 & 0.813                                 &                                         & 0.813                                &                                         &                                   \\ \hline
4                                    & Simple Logistic                         & 80.21                                  & 0.802                                 & 0.816                                 & 0.660                                   & 0.802                                & 0.724                                   & 0.051                             \\ \hline
5                                    & SMO                                     & 81.31                                  & 0.813                                 & 0.813                                 &                                         & 0.813                                &                                         &                                   \\ \hline
6                                    & Voted Perceptron                        & 81.31                                  & 0.813                                 & 0.813                                 &                                         & 0.813                                &                                         &                                   \\ \hline
7                                    & Random Forest                           & 82.41                                  & 0.824                                 & 0.471                                 & 0.813                                   & 0.824                                & 0.817                                   & 0.382                             \\ \hline
8                                    & Adaboost                                & 79.67                                  & 0.797                                 & 0.794                                 & 0.699                                   & 0.797                                & 0.730                                   & 0.006                             \\ \hline
9                                    & Random Subspace                         & 81.31                                  & 0.813                                 & 0.813                                 &                                         & 0.813                                &                                         &                                   \\ \hline
10                                   & Bayes Net                               & 81.31                                  & 0.813                                 & 0.813                                 &                                         & 0.813                                &                                         &                                   \\ \hline
11                                   & Naive Bayes                             & 77.47                                  & 0.775                                 & 0.618                                 & 0.751                                   & 0.775                                & 0.761                                   & 0.177                             \\ \hline
12                                   & LWL                                     & 81.31                                  & 0.813                                 & 0.813                                 &                                         & 0.813                                &                                         &                                   \\ \hline
\end{tabular}
\end{table*}

\subsection{Results using Online Classifiers}

Having established that the generated features from the data can be used to train a learning model for the task of dengue classification, the focus on this section is to evaluate its viability in an online learning setting. To this end, we conduct two sets of experiments, discussed in the following subsections.

\subsubsection{Experiments on Incremental Classifier Training}

The first set of experiments aims to establish the robustness in training classifiers in an online setting using our dataset. We use the simple perceptron algorithm for this evaluation. We begin with a small sized training set used to train the perceptron model. We incrementally increase the number of samples in the training dataset and compute the error rate, number of updates required for the learning model parameters to converge, and the computational time in training the model. This is done for two cases. In the first case, the classifier model is re-trained with number of samples added incrementally, as well as the samples from which it was trained in the previous iteration. The results in this setting are shown in Table \ref{res:percep1}.

\begin{table*}[hbtp]
\centering
\caption{Algorithm :- Perceptron used to examine influence by increasing samples  in training dataset on mistake rate, number of updates and execution time by retraining the model using the entire training dataset}
\label{res:percep1}
\scalebox{1.0}{
\begin{tabular}{|c|c|l|l|l|}
\hline
\multicolumn{1}{|l|}{\textbf{S.No.}} & \textbf{\# Records} & \multicolumn{1}{c|}{\textbf{Error Rate}} & \multicolumn{1}{c|}{\textbf{\# Updates}} & \multicolumn{1}{c|}{\textbf{CPU Time(s)}} \\ \hline
1                                    & 19                          & 0.3684 $\pm$ 0.0783                          & 7.00 $\pm$ 1.49                                & 0.0038 $\pm$ 0.0023                          \\ \hline
2                                    & 37                          & 0.3216 $\pm$ 0.0547                          & 11.90 $\pm$ 2.02                               & 0.0058 $\pm$ 0.0018                          \\ \hline
3                                    & 55                          & 0.3082 $\pm$ 0.0331                          & 16.95 $\pm$ 1.82                               & 0.0077 $\pm$ 0.0021                          \\ \hline
4                                    & 73                          & 0.3027 $\pm$ 0.0301                          & 22.10 $\pm$ 2.20                               & 0.0083 $\pm$ 0.0016                          \\ \hline
5                                    & 91                          & 0.3077 $\pm$ 0.0323                          & 28.00 $\pm$  2.94                              & 0.0106 $\pm$ 0.0018                          \\ \hline
6                                    & 110                         & 0.3018 $\pm$ 0.0271                          & 33.20 $\pm$ 2.98                                 & 0.0129 $\pm$ 0.0031                          \\ \hline
7                                    & 128                         & 0.3066 $\pm$ 0.0272                          & 39.25 $\pm$ 3.48                               & 0.0157 $\pm$ 0.0017                          \\ \hline
8                                    & 146                         & 0.3034 $\pm$ 0.0217                          & 44.30 $\pm$ 3.16                               & 0.0167 $\pm$ 0.0026                          \\ \hline
9                                    & 164                         & 0.2951 $\pm$ 0.0237                          & 48.40 $\pm$ 3.89                               & 0.0202 $\pm$ 0.0029                          \\ \hline
10                                   & 182                         & 0.2981 $\pm$ 0.0201                          & 54.25 $\pm$ 3.65                               & 0.0221 $\pm$ 0.0027                          \\ \hline
\end{tabular}%
}
\end{table*}

In the second case, the classifier model is trained using only the new samples which are incrementally added at each iteration. These results are shown in Table \ref{res:percep2}. It can be seen that error rates and number of updates required are comparable in both the cases. The error reate decreases with an increase in training data. It indicates that the classifier model is robustly trained for dengue detection.

\begin{table*}[hbtp]
\centering
\caption{Algorithm :- Perceptron used to examine influence by incrementally increasing samples in training dataset on mistake rate, number of updates and execution time by incrementally training the learning model}
\label{res:percep2}
\scalebox{1.0}{
\begin{tabular}{|c|c|l|l|l|}
\hline
\textbf{S.No.} & \textbf{\# Records} & \multicolumn{1}{c|}{\textbf{Error Rate}} & \multicolumn{1}{c|}{\textbf{\# Updates}} & \multicolumn{1}{c|}{\textbf{CPU Time (s)}} \\ \hline
1              & 19                          & 0.3684 $\pm$ 0.0783                          & 7.00 $\pm$ 1.49                                & 0.0021 $\pm$ 0.0002                           \\ \hline
2              & 37                          & 0.3216 $\pm$ 0.0547                          & 11.90 $\pm$ 2.02                               & 0.0049 $\pm$ 0.0015                          \\ \hline
3              & 55                          & 0.3082 $\pm$ 0.0331                          & 16.95 $\pm$ 1.82                               & 0.0072 $\pm$ 0.0020                          \\ \hline
4              & 73                          & 0.3027 $\pm$ 0.0301                          & 22.10 $\pm$ 2.20                               & 0.0101 $\pm$ 0.0086                          \\ \hline
5              & 91                          & 0.3077 $\pm$ 0.0323                          & 28.00 $\pm$  2.94                              & 0.0112 $\pm$ 0.0014                          \\ \hline
6              & 110                         & 0.3018 $\pm$ 0.0271                          & 33.20 $\pm$ 2.98                                 & 0.0126 $\pm$ 0.0022                          \\ \hline
7              & 128                         & 0.3066 $\pm$ 0.0272                          & 39.25 $\pm$ 3.48                               & 0.0159 $\pm$ 0.0023                           \\ \hline
8              & 146                         & 0.3034 $\pm$ 0.0217                          & 44.30 $\pm$ 3.16                               & 0.0178 $\pm$ 0.0018                          \\ \hline
9              & 164                         & 0.2951 $\pm$ 0.0237                          & 48.40 $\pm$ 3.89                               & 0.0192 $\pm$ 0.0027                          \\ \hline
10             & 182                         & 0.2981 $\pm$ 0.0201                          & 54.25 $\pm$ 3.65                               & 0.0207 $\pm$ 0.0025                          \\ \hline
\end{tabular}%
}
\end{table*}

One may also note that the CPU time is lower for the case when the classifier model is trained in a purely incremental fashion, i.e., it is re-trained using only the new samples rather that the entire dataset. As the error rates and number of updates are comparable for both the experimental settings, choosing an approach to train the classifier in an incremental manner is beneficial as it saves on computational time, which is critical when the dengue detection system is deployed in a real-time system.

\subsubsection{Results using various classifier models}

The results using multiple online classification models trained incrementally are shown in Table \ref{res:online}. It can be seen that the lowest error rate is obtained for AROW algorithm, while the number of updates required are least for the IELLIP algorithm. This indicates that the generalization of AROW is the best for this dataset, while computational cost is least for IELLIP. The CPU time involved in training these online models is also shown. Though the CPU time is not a bottleneck in the present scenario since the dataset size is limited, the low CPU times indicate that these models are viable for implementation in practical systems where they can be used for real-time dengue disease classification, without heavy computing costs and obtaining predictions within reasonable compute times.

\begin{table*}[hbtp]
\centering
\caption{Results using online classifiers for dengue detection}
\label{res:online}
\scalebox{1.0}{
\begin{tabular}{|l|l|l|l|l|}
\hline
\multicolumn{1}{|c|}{S.No.} & \multicolumn{1}{c|}{Algorithm} & \multicolumn{1}{c|}{Error Rate} & \multicolumn{1}{c|}{No. of Updates} & \multicolumn{1}{c|}{CPU Time (s)} \\ \hline
1                           & AROW                           & \textbf{0.1901 $\pm$ 0.0055}      & 175.45$\pm$ 5.30                      & 0.0411 $\pm$ 0.0026                 \\ \hline
2                           & OGD                            & 0.1934 $\pm$ 0.0088               & 78.75 $\pm$ 3.19                      & 0.0288 $\pm$ 0.0015                 \\ \hline
3                           & SCW2                           & 0.1973 $\pm$ 0.0175               & 130.90 $\pm$14.31                     & 0.0355 $\pm$ 0.0036                 \\ \hline
4                           & NHERD                          & 0.1995$\pm$ 0.0288                & 175.00$\pm$4.65                       & 0.0363 $\pm$ 0.0023                 \\ \hline
5                           & NAROW                          & 0.2239 $\pm$ 0.0709               & 177.55$\pm$5.60                       & 0.0434$\pm$ 0.0010                  \\ \hline
6                           & SCW                            & 0.2272$\pm$ 0.1307                & 65.40 $\pm$ 38.90                     & 0.0311 $\pm$ 0.0027                 \\ \hline
7                           & PA                             & 0.2953 $\pm$ 0.0267               & 93.25 $\pm$ 7.68                      & 0.0243 $\pm$ 0.0026                 \\ \hline
8                           & PA2                            & 0.2953 $\pm$ 0.0267               & 93.90 $\pm$ 7.83                      & 0.0266 $\pm$ 0.0023                 \\ \hline
9                           & PA1                            & 0.2963 $\pm$ 0.0267               & 93.25 $\pm$ 7.68                      & 0.0254 $\pm$ 0.0019                 \\ \hline
10                          & Perceptron                     & 0.2964 $\pm$ 0.0271               & 53.95 $\pm$ 4.93                      & 0.0206 $\pm$ 0.0031                 \\ \hline
11                          & IELLIP                         & 0.2967$\pm$ 0.0266                & \textbf{54.00$\pm$ 4.84}                       & 0.0297 $\pm$ 0.0018                 \\ \hline
12                          & ALMA                           & 0.2981 $\pm$ 0.0265               & 55.75 $\pm$ 4.96                      & 0.0253 $\pm$ 0.0017                 \\ \hline
13                          & CW                             & 0.2989 $\pm$ 0.0264               & 110.35 $\pm$ 5.83                     & 0.0326 $\pm$ 0.0016                 \\ \hline
14                          & SOP                            & 0.3003$\pm$ 0.0200                & 54.65 $\pm$ 3.65                      & 0.0337 $\pm$ 0.0022                 \\ \hline
15                          & aROMMA                         & 0.0323 $\pm$ 0.0449               & 62.05 $\pm$ 8.37                      & 0.0230 $\pm$ 0.0018                 \\ \hline
16                          & ROMMA                          & 0.3343 $\pm$ 0.0481               & 60.85 $\pm$ 8.76                      & 0.0223 $\pm$ 0.0021                 \\ \hline
\end{tabular}%
}
\end{table*}

\section{Conclusions and Future Work\label{sec:conclusions}}
This paper presented the design architecture and comprehensive evaluation of an online dengue disease prediction model that can be incorporated in a healthcare information system for flagging patients with high probability of being diagnosed with dengue. The model uses simple features which are derived using a combination of recorded symptoms and results of laboratory investigations relevant for diagnosing dengue. The learning model has been trained in an online manner which allows it to incrementally generalize better as more labeled training data is validated by the clinician. Results demonstrate the accuracy and feasibility of the proposed model being deployed in a practical setting for assisting clinicians by providing a data-driven decision model, thereby contributing to overall improvement in healthcare services for disease detection, as well as in identifying trends in varying symptoms for such diseases. Future work involves expanding of the feature space used in this model to include geographic and demographic information, as well as other parameters such as patient vitals. Similar models can also be developed for detection of other diseases that are of emerging interest to the healthcare practitioners, public health researchers and healthcare policy makers.

\bibliographystyle{IEEEtran}
\bibliography{references}

\end{document}